\documentclass[10pt,twocolumn,letterpaper]{article}

\usepackage[pagenumbers]{cvpr} % To force page numbers, e.g. for an arXiv version
\usepackage{CJKutf8}      % UTF-8 のまま日本語を入力

\usepackage{makecell}  % in your preamble

\definecolor{cvprblue}{rgb}{0.21,0.49,0.74}
\usepackage[pagebackref,breaklinks,colorlinks,allcolors=cvprblue]{hyperref}

\def\paperID{10} % *** Enter the Paper ID here
\def\confName{CVPR}
\def\confYear{2025}

\title{CAG‐VLM : Fine tuning of a large-scale model to recognize angiographic images for next-generation diagnostic systems}

\author{%
  Yuto Nakamura\textsuperscript{1,2}\quad
  Satoshi Kodera\textsuperscript{2} \quad
  Haruki Settai\textsuperscript{1,2}\quad
  Hiroki Shinohara\textsuperscript{2}\quad
  Masatsugu Tamura\textsuperscript{2}\\[0.8ex]
  Tomohiro Noguchi\textsuperscript{2}\quad
  Tatsuki Furusawa\textsuperscript{2}\quad
  Ryo Takizawa\textsuperscript{1,2}\quad
  Tempei Kabayama\textsuperscript{1,2}\quad
  Norihiko Takeda\textsuperscript{2}\\[1.5ex]
  {\small
    \textsuperscript{1}The University of Tokyo\quad
    \textsuperscript{2}The University of Tokyo Hospital
  }
}

\begin{document}
\maketitle
\begin{abstract}
Coronary angiography (CAG) is the gold‐standard imaging modality for evaluating coronary artery disease, but its interpretation and subsequent treatment planning rely heavily on expert cardiologists. To enable AI‐based decision support, we introduce a two‐stage, physician‐curated pipeline and a bilingual (Japanese/English) CAG image–report dataset. First, we sample 6,597 frames from 210 exams and annotate them for key‐frame detection and left/right laterality; a ConvNeXt‐Base CNN trained on this data achieves 0.96 F1 on laterality classification, even on low‐contrast frames. Second, we apply the CNN to 243 independent exams, extract 1,101 key frames, and pair each with its pre‐procedure report and expert‐validated diagnostic and treatment summary, yielding a parallel corpus. We then fine‐tune three open‐source VLMs (PaliGemma2, Gemma3, and ConceptCLIP‐enhanced Gemma3) via LoRA and evaluate them using VLScore and cardiologist review. Although PaliGemma2 w/LoRA attains the highest VLScore, Gemma3 w/LoRA achieves the top clinician rating (mean 7.20/10); we designate this best‐performing model as CAG‐VLM. These results demonstrate that specialized, fine‐tuned VLMs can effectively assist cardiologists in generating clinical reports and treatment recommendations from CAG images.
\end{abstract}
    
\section{Introduction}
\label{sec:intro}
Coronary artery disease (CAD) is a leading cause of morbidity and mortality worldwide \cite{James2018}, and coronary angiography (CAG) is widely employed as the gold standard for its diagnosis \cite{Vrints2024ESC}. In CAG, a contrast agent is injected into the coronary vessels and X-ray imaging is used to directly assess the presence and severity of stenoses or occlusions. Typically, each exam comprises multiple videos (lasting several to ten seconds) of both the left and right coronary arteries from various viewing angles. Physicians must review all these cine sequences to formulate their findings and decide on an optimal treatment plan. Accurately interpreting monochrome angiograms thus requires extensive expertise and considerable effort \cite{Shivaie2024Interobserver}\cite{Leape2000-hk}, driving strong demand for automated support \cite{Riggs2025Assessing}.

\begin{figure}[t]
  \centering
  \includegraphics[width=\linewidth]{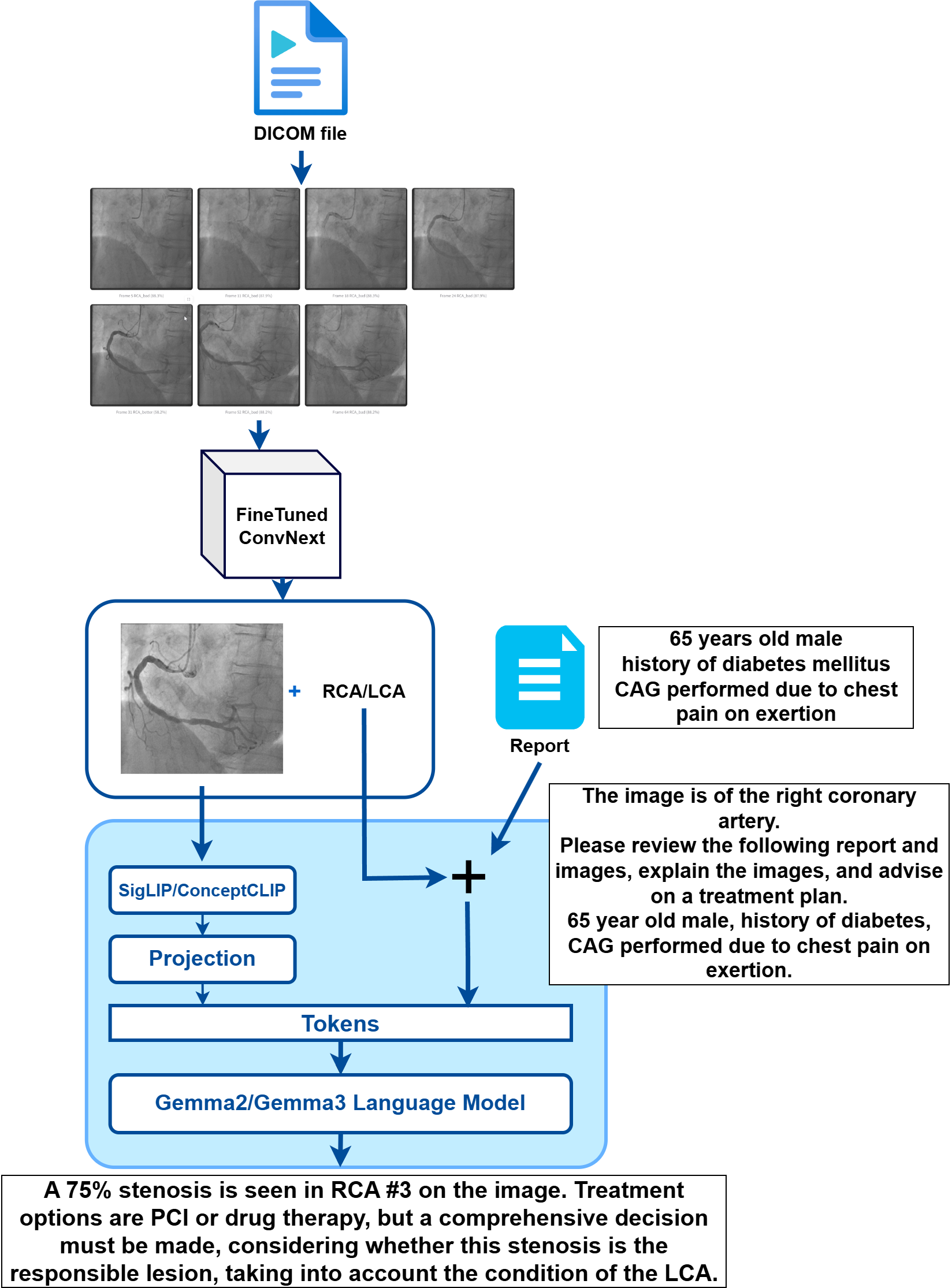}
  \caption{Overview of our two‐stage pipeline.}
  \label{fig:pipeline}
\end{figure}

Recent advances in vision-language models (VLMs) have spurred efforts to fine-tune foundation models with medical data, as exemplified by systems like Med-Gemini\cite{MedGemini} and Llava-Med\cite{LLaVA-Med}. Moreover, state-of-the-art foundation models have begun to exhibit a certain degree of medical knowledge. For instance, PaliGemma2 has been reported to generate appropriate diagnostic reports for X-ray images\cite{PaliGemma2}. The emergence of such open-source multi-modal models with embedded medical expertise suggests new possibilities for AI-driven diagnosis of coronary artery conditions.

Deep learning–based stenosis detection\cite{CathAI}\cite{DeepCoro} and Vision-Language Models (VLMs) for report generation in CT and X-ray domains have seen rapid progress\cite{JRadiEvo}\cite{PaliGemma2}. However, two critical challenges remain unaddressed for invasive CAG:

\begin{enumerate}
  \item \textbf{Data scarcity} \\
    No public dataset links CAG frames with patients’ clinical histories and expert-validated diagnostic plus treatment summaries.
  \item \textbf{Delayed VLM adoption} \\
    Compared to noninvasive modalities (CT, X-ray), there are virtually no applications of large-scale multi-modal VLMs to CAG, leaving their potential largely unexplored.
\end{enumerate}

To overcome these gaps, we propose a physician‐curated, two‐phase pipeline. First, we sampled 6,597 frames from 210 CAG exams and annotated each for laterality (LCA/RCA) and key-frame category (better/bad/others). A ConvNeXt‐based CNN fine-tuned on this data achieves automatic key-frame extraction and label assignment. Second, we applied this CNN to 243 independent exams to extract 1,101 frames, pairing each with the patient’s pre-examination report and expert-validated diagnostic and treatment plan summaries. We then fine-tuned three VLM variants—PaliGemma2, Gemma3, and a ConceptCLIP‐enhanced Gemma3—via LoRA. Finally, quantitative evaluation using the VLSCore\cite{VLScore} framework and qualitative assessment by cardiologists confirm that our fine-tuned models surpass both closed-source baselines and their own pre-fine-tuned versions in clinical expressiveness.

\section{Related Work}
\label{sec:relatedwork}

\subsection{Vision-Language Models for Medical Imaging}
Vision–Language Models (VLMs) have seen rapid advances in medical image analysis. Early work such as CheXbert combined hand‐crafted and auto‐generated rules to extract findings from chest X-ray reports\cite{CheXbert}, while RadGraph automatically identifies clinical entities and their relationships within radiology reports\cite{RadGraph}. More recently, large‐scale foundation models like Llava-Med\cite{LLaVA-Med} and Med-Gemini\cite{MedGemini} have been fine-tuned across multiple healthcare tasks, demonstrating versatility beyond single‐purpose applications.

Pretrained multimodal models have also emerged: PMC-CLIP leverages millions of image–caption pairs from PubMed Central to boost performance on medical image retrieval and classification benchmarks\cite{PMC-CLIP}; BiomedCLIP \cite{BiomedCLIP}, MedCLIP\cite{MedCLIP}, and ConceptCLIP\cite{ConceptCLIP} similarly learn from massive biomedical image–text corpora to achieve state-of-the-art results on chest X-ray, MRI, and pathology tasks. In contrast to these embedding-based approaches, generative VLMs such as PaliGemma2 employ an encoder–decoder transformer to produce full-length radiology reports; when fine-tuned on the MIMIC-CXR dataset, PaliGemma2 achieved a RadGraph F1 score of 29.5\%, outperforming prior methods in both report quality and clinical consistency\cite{PaliGemma2}.  However, these efforts have focused almost exclusively on noninvasive modalities (CT, X-ray, MRI), and applications of VLMs to invasive procedures such as coronary angiography remain largely unexplored.

\subsection{Deep Learning for Coronary Angiography: Analysis \& Datasets}
Deep learning in coronary angiography (CAG) has concentrated on stenosis detection, laterality classification, and vessel segmentation using CNN-based architectures. \citet{Eschen} achieved a laterality classification F1 of 0.99 on 3,500 CAG videos with a 3D-CNN\cite{Eschen}. CathAI integrates view selection, laterality labeling, stenosis localization, and severity estimation into a single pipeline, delivering high quantification and localization accuracy\cite{CathAI}. Cong et al. combined CNN + LSTM for key-frame selection and FPN for multi-class stenosis classification (mild, moderate, occlusion), reaching an AUC of 0.86 and sensitivity of 0.96\cite{Cong2023-yx}.More recently, DeepCoro\cite{DeepCoro}introduced an eight-stage, video-based analysis pipeline trained on 182,418 CAG cine loops, achieving an AUC of approximately 0.83 for significant stenosis detection—surpassing earlier methods in end-to-end clinical performance\cite{DeepCoro}.

Several CAG-specific datasets have been released for benchmarking deep learning algorithms. ARCADE comprises two 1,200-image cohorts—one for vessel segmentation and one for stenosis detection—each split into 1,000 training images annotated in COCO format and 200 validation images; the segmentation cohort labels 26 coronary regions according to the SYNTAX score methodology, while the stenosis cohort provides atherosclerotic plaque annotations\cite{ARCADE}. CoronaryDominance includes 1,574 invasive angiography videos labeled for right- versus left-dominant coronary circulation under varied image-quality conditions, enabling evaluation of domain generalization\cite{CoronaryDominance}. CADICA contains 42 CAG cases with expert-reviewed stenosis annotations\cite{CADICA}. 

While these resources are invaluable, they face challenges: domain shifts from different imaging systems, potential data leakage when adjacent frames are highly similar, and predominantly English reports that limit non-English clinical contexts. In contrast, our work builds both a laterality/key-frame detection set and a Japanese multimodal corpus pairing pre-procedure reports with expert‐validated diagnostic and treatment summaries for 1,101 key frames, addressing these limitations.
\citet{Eschen}
\section{Method}

\subsection{Data Preparation}

To train our key‐frame extractor and laterality classifier, we sampled 707 DICOM cine sequences from 210 CAG examinations and identified 6,597 still frames at local extrema of pixel‐intensity mean and variance, enforcing a minimum 5‐frame gap to eliminate near‐duplicates. Each frame was labeled into one of six combined laterality–keyframe categories (\texttt{LCA\_better}, \texttt{LCA\_bad}, \texttt{LCA\_other}, \texttt{RCA\_better}, \texttt{RCA\_bad}, \texttt{RCA\_other}), with cross‐checking by two physicians for consistency (Table~\ref{tab:cnn-split}).

\begin{figure}[t]
  \centering
  \begin{subfigure}[b]{0.30\linewidth}
    \includegraphics[width=\linewidth]{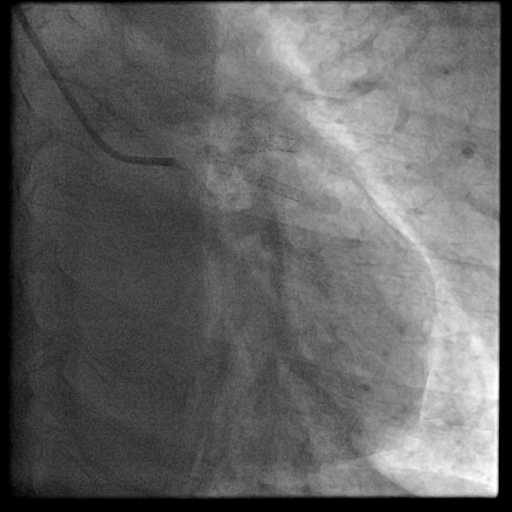}
    \caption{LCA\_bad}
  \end{subfigure}
  \begin{subfigure}[b]{0.30\linewidth}
    \includegraphics[width=\linewidth]{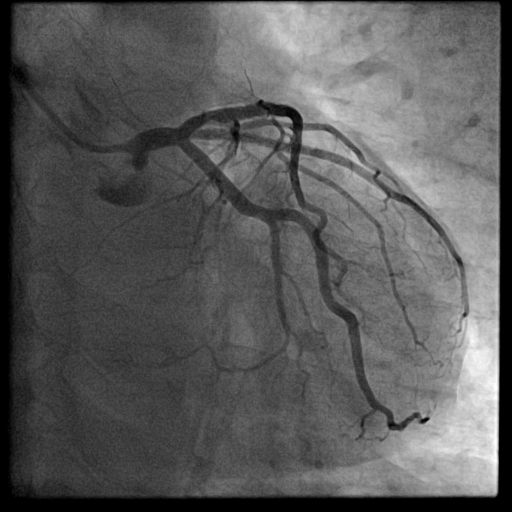}
    \caption{LCA\_better}
  \end{subfigure}
  \begin{subfigure}[b]{0.30\linewidth}
    \includegraphics[width=\linewidth]{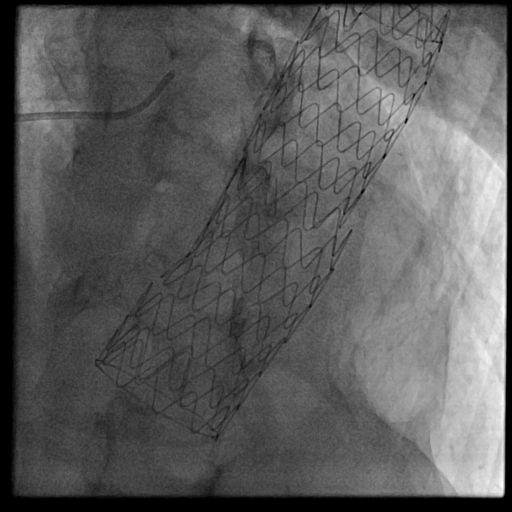}
    \caption{LCA\_others}
  \end{subfigure}

  \vspace{0.5em}

  \begin{subfigure}[b]{0.30\linewidth}
    \includegraphics[width=\linewidth]{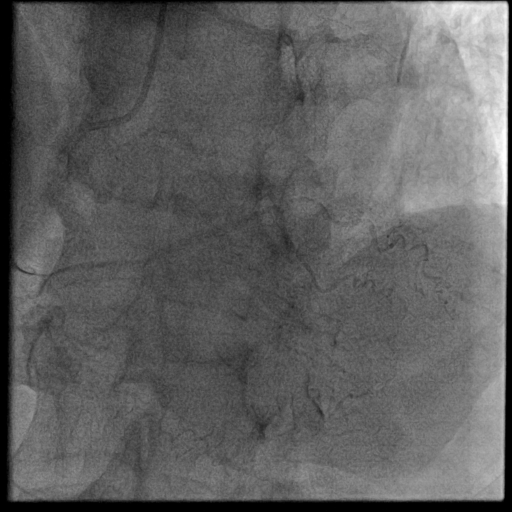}
    \caption{RCA\_bad}
  \end{subfigure}
  \begin{subfigure}[b]{0.30\linewidth}
    \includegraphics[width=\linewidth]{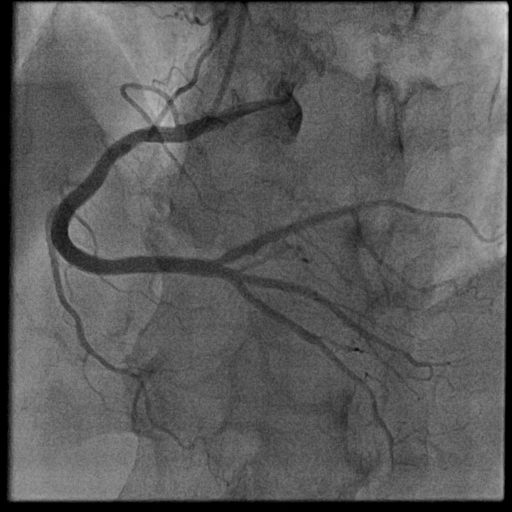}
    \caption{RCA\_better}
  \end{subfigure}
  \begin{subfigure}[b]{0.30\linewidth}
    \includegraphics[width=\linewidth]{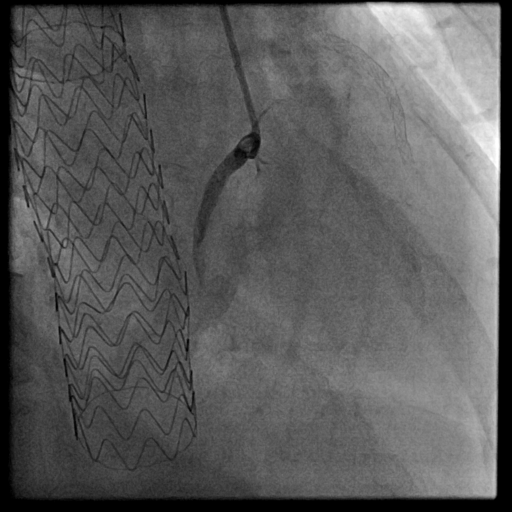}
    \caption{RCA\_others}
  \end{subfigure}

  \caption{Examples of the six CNN training classes.}
  \label{fig:cnn-class-examples}
\end{figure}

Applying the trained CNN to an independent set of 243 examinations, we selected the 1,101 highest‐confidence frames from 530 videos. Each key frame was paired with its original Japanese pre‐procedure clinical report (patient history and indication) and an expert‐validated diagnostic and treatment summary. To enable quantitative VLScore evaluation, we also created an English‐translated version of every report, yielding two parallel corpora (Japanese and English).  

Table~\ref{tab:vlm-split} summarizes how these 530 videos (1,101 images) were divided into non‐overlapping training, validation, and test splits, in terms of cases (exams), videos, and key frames.

\begin{figure*}[t]
  \centering
  % ----- (a) 画像サブフィギュア -----
  \begin{subfigure}[b]{0.5\linewidth}
    \includegraphics[width=\linewidth]{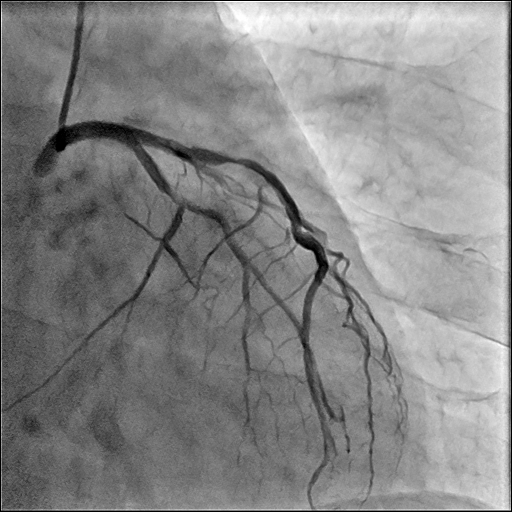}
    \caption{Left coronary angiogram (LCA).}
    \label{fig:example-angio}
  \end{subfigure}
  \hfill
  % ----- (b) テキスト側サブフィギュア -----
  \begin{subfigure}[b]{0.45\linewidth}
    \small
    \textbf{Prompt:}\\
    “This image is a left coronary angiogram. Based on the image and the clinical report below, provide an explanation and advice on future management.”\\[0.5ex]
    \textbf{Report (JP):}\\
    \begin{CJK}{UTF8}{ipxm}
    84歳男性．2021年2月10日に冠動脈造影（CAG）施行。\\[0.5ex]
    \end{CJK}
    \textbf{Report (EN):}\\
    84‐year‐old male. CAG performed 2021/2/10.\\[0.5ex]
    \textbf{Ground Truth (JP):}\\
    \begin{CJK}{UTF8}{ipxm}
    LCX \#11に75\%の狭窄が確認できます。他のアンギオ画像や、合併症を考慮して、治療方針を検討してください。
    \end{CJK}\\[0.5ex]
    \textbf{Ground Truth (EN):}\\
    LCX \#11 shows 75\% stenosis. Consider treatment plan in context of other angiographic findings and comorbidities.\\[0.5ex]
    \textbf{PaliGemma2 w/LoRA:}\\
    LAD septal shows 90\% stenosis. Recommend watchful waiting without medical therapy.\\[0.5ex]
    \textbf{Gemma3 w/LoRA:}\\
    LCX \#12 shows 90\% stenosis. Consider treatment plan in context of other angiographic findings and comorbidities.\\[0.5ex]
    \textbf{ConceptCLIP‐Gemma3 w/LoRA:}\\
    LCX \#11 shows 90\% stenosis. Consider treatment plan in context of other angiographic findings and comorbidities.\\[0.5ex]
    \textbf{Comment:}\\
    PaliGemma2 mislabels LAD septal and inappropriately recommends no therapy. Gemma3 confuses \#12/11—adjacent vessels prone to mix‐up. ConceptCLIP‐Gemma3 correctly identifies LCX \#11.
    \caption{Prompt, Japanese \& English clinical report, ground truth annotation, and outputs of three fine-tuned VLMs with expert commentary.}
    \label{fig:example-angio-text}
  \end{subfigure}

  \caption{Multimodal Input and Model Outputs for Left Coronary Angiography. 
    (a) Left coronary angiogram (LCA). 
    (b) Prompt, clinical report in Japanese \& English, ground truth stenosis annotation, and outputs of three fine-tuned vision–language models with expert commentary.}
  \label{fig:vlm-output-example}
\end{figure*}

Figure~\ref{fig:vlm-output-example} illustrates one such paired instance—showing the extracted angiogram frame, the original Japanese report, its English translation, and the expert summary—demonstrating how our bilingual dataset is structured.

\subsection{Model Architectures}

Our key‐frame extraction network is based on the ConvNeXt–Base backbone. We replaced the original 1,000‐way head with a single six‐way classification head that directly predicts one of the combined laterality–keyframe classes.

For diagnostic narrative generation, we employ three open‐source VLM variants: PaliGemma2, Gemma3, and ConceptCLIP‐Gemma3.

To better capture medical‐image–specific features, we replaced Gemma3’s original Vision Encoder with ConceptCLIP—a multimodal model fine‐tuned on medical data and reported to achieve state‐of‐the‐art performance. We hypothesized that leveraging ConceptCLIP’s medical knowledge would improve downstream report generation accuracy.

All three backbones were then adapted with Low‐Rank Adaptation (LoRA) modules injected into each transformer’s projection matrices (\texttt{q\_proj}, \texttt{k\_proj}, \texttt{v\_proj}, \texttt{o\_proj}, \texttt{gate\_proj}, \texttt{up\_proj}, \texttt{down\_proj}). In addition, the cross‐modal projection layer that maps visual embeddings to the language model’s token embeddings was also fine‐tuned, ensuring the vision–language interface is jointly optimized. We used a rank (and corresponding $\alpha$) of 16 for PaliGemma2 and 8 for both Gemma3 and ConceptCLIP‐Gemma3 to minimize additional parameters.

\subsection{Training Procedures}

\paragraph{CNN Training}
The ConvNeXt–Base network was trained for 200 epochs on two A6000 Ada GPUs using AdamW (learning rate $1\times10^{-4}$, batch size 128). We optimized a standard cross‐entropy loss over the six classes and applied TrivialAugmentWide for data augmentation.To prevent overlap of images from the same video across splits, each video’s images were assigned to a single split. The resulting data partitioning is shown in Table\ref{tab:cnn-split}.

\begin{table}
  \centering
  \caption{\label{tab:cnn-split}CNN training dataset split ensuring no overlap of images from the same video.}
  \begin{tabular}{lrrr}
    \toprule
    Split & Cases & Videos & Images \\
    \midrule
    train & 116 & 565 & 5260 \\
    val   &  46 &  70  &  657 \\
    test  &  48 &  72  &  680 \\
    \bottomrule
  \end{tabular}
\end{table}

\paragraph{VLM Fine‐Tuning}
For each VLM backbone, we fine‐tuned two separate models: one on the Japanese corpus and one on the English corpus. Exams were split at the patient level to prevent leakage (Table~\ref{tab:vlm-split}). Training ran for 5 epochs on four H100 GPUs (AdamW, LR = 1\,$\times$\,10$^{-5}$), with LoRA modules as specified.

\begin{table}
  \centering
  \caption{\label{tab:vlm-split}VLM fine‐tuning dataset split (exams/patients non‐overlapping).}
  \begin{tabular}{lrrr}
    \toprule
    Split & Cases & Videos & Images \\
    \midrule
    train & 194 & 425 & 889 \\
    val   &  24 &  49  & 105 \\
    test  &  25 &  56  & 107 \\
    \bottomrule
  \end{tabular}
\end{table}

This pipeline—from carefully curated data preparation through architecture design and standardized training—establishes a robust foundation for automated CAG report generation and treatment planning support.
\section{Experiments}
\subsection{Experimental Setup}

In this study, we separately evaluate the Key–Frame \& Laterality CNN and VLMs. The CNN is assessed on its six‐class classification (laterality + highlight) performance using F1 score.

For VLMs, we adopt the embedding‐based VLScore metric. VLScore quantifies the area of the triangle formed by the image embedding \(i_e\), the generated report embedding \(r_e\), and the ground‐truth report embedding \(g_e\), computing a distance \(T\) as:

\begin{equation}
T(i,g,r)
= \frac12
  \sqrt{%
    \begin{aligned}
      \langle i_e - g_e,\,i_e - g_e\rangle
      \,\langle i_e - r_e,\,i_e - r_e\rangle \\
      \quad -\,\langle i_e - g_e,\,i_e - r_e\rangle^2
    \end{aligned}
  }
\label{eq:dist}
\end{equation}

\begin{equation}
    \mathrm{VLScore}(i, g, r)
    = \max\!\Bigl\{1 - \tfrac{T(i, g, r)}{C},\,0\Bigr\}
\label{eq:VLS}
\end{equation}

where \(C\) is the theoretical maximum area of the triangle. A smaller \(T\) indicates that the three embeddings are closer in representation space. Unlike text‐only metrics such as BLEU or ROUGE, VLScore captures both visual context and clinical terminology alignment.

In our experiments, we use SigLIP2-so400m, MedCLIP, and ConceptCLIP as embedding models, and compare un‐fine‐tuned OpenAI o4-mini, PaliGemma2, and Gemma3 as generation backbones. Outputs are generated with few‐shot prompts matching the dataset format, and VLScore is computed according to Eqs.~(\ref{eq:dist})–(\ref{eq:VLS}).

These metrics and procedures allow us to quantitatively compare the CNN’s key‐frame extraction accuracy and the VLMs’ clinical expressiveness.

\subsection{Key-Frame \& Laterality Model Results}

\begin{figure}[t]
  \centering
  \includegraphics[width=\linewidth]{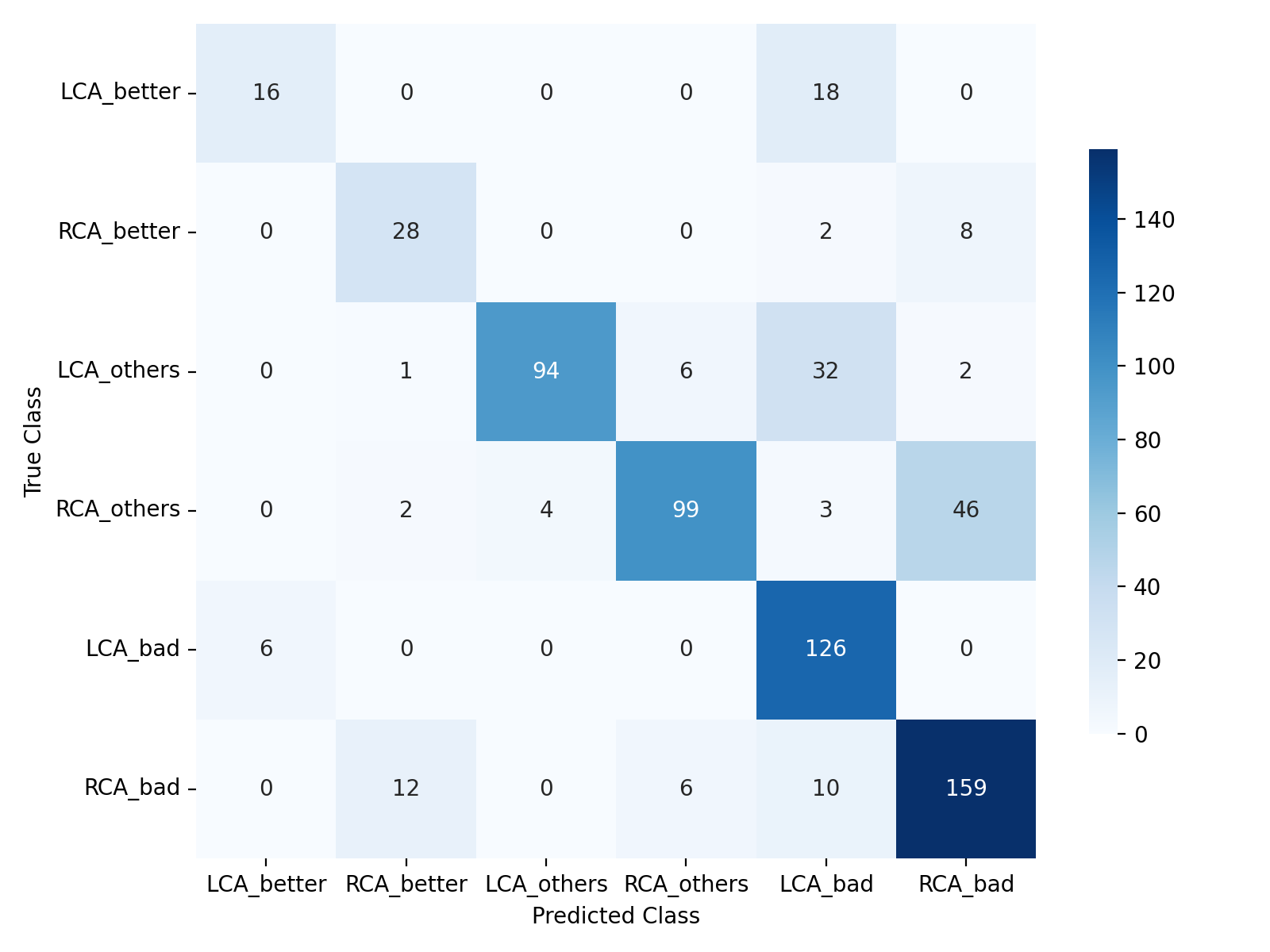}
  \caption{Confusion matrix of the CNN on the test set.}
  \label{fig:cnn-confusion}
\end{figure}

The classification performance of the CNN model on the test set is visualized by the confusion matrix in Figure \ref{fig:cnn-confusion}. The overall weighted average F1 score across all six classes is 0.766. When considering only laterality, the F1 score rises to 0.959, indicating that left–right discrimination remains highly accurate even on low‐contrast or degraded frames. Treating key‐frame extraction as a three‐class task (LCA\_better, RCA\_better, others) yields a weighted average F1 score of 0.925, demonstrating excellent precision in selecting diagnostically informative frames. Common confusions—such as between “LCA\_better” and “LCA\_bad” or “RCA\_better” and “RCA\_bad”—originate from boundary ambiguities during annotation and are unlikely to impact practical usage. These results confirm that the CNN reliably extracts key frames and assigns laterality, producing robust inputs for subsequent VLM training.

\subsection{VLM Quantitative Results}
\begin{figure*}[t]
  \centering
  \includegraphics[width=0.9\linewidth]{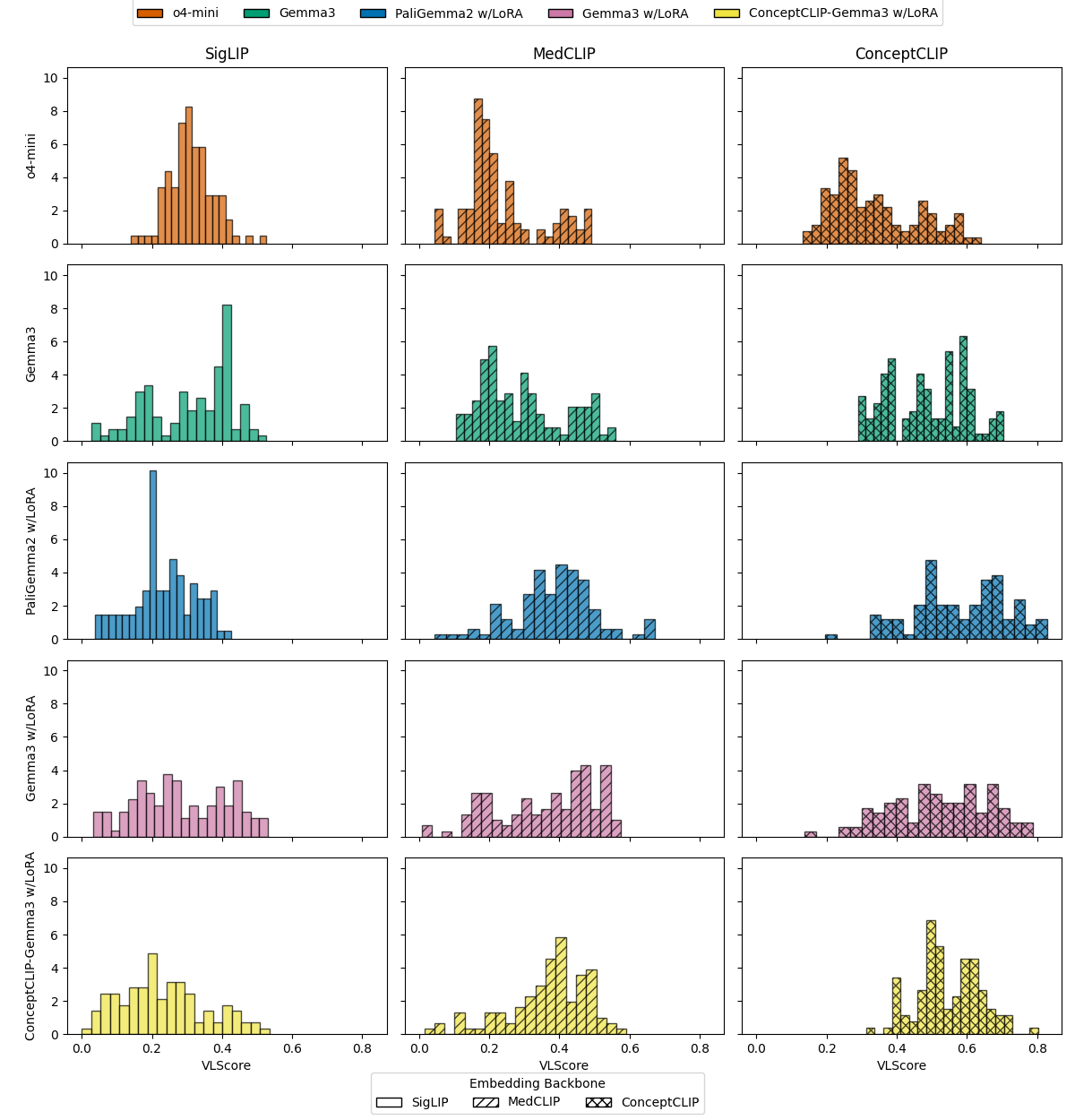}
  \caption{VLScore distributions for each model across three embedding backbones (SigLIP2‐so400m, MedCLIP, ConceptCLIP).}
  \label{fig:VLSCore_histogram}
\end{figure*}

The fine-tuned VLMs—each trained on the English side of our newly constructed English-Japanese parallel CAG dataset—were evaluated with VLScore using three embedding backbones: SigLIP2-so400m, MedCLIP, and ConceptCLIP. In Figure \ref{fig:VLSCore_histogram}, each column presents results computed with the same embedding backbone, while each row corresponds to a different VLM. Scores that fall farther to the right denote a higher VLScore—i.e., a smaller geometric area formed by the embeddings of the generated text, the ground-truth text, and the input image—indicating better multimodal alignment. The unfine-tuned PaliGemma2 is omitted because it produced only generic responses such as “Sorry, as a base VLM I am not trained to answer this question.”

Table~\ref{tab:vlscore-all} summarizes mean and standard deviation of VLScore for each model under each embedding.

\begin{table*}[t]
  \centering
  \caption{VLScore mean and standard deviation for each model across three embedding backbones.}
  \label{tab:vlscore-all}
  \begin{tabular}{lcccccc}
    \toprule
    Model                        & \multicolumn{2}{c}{SigLIP2‐so400m} & \multicolumn{2}{c}{MedCLIP} & \multicolumn{2}{c}{ConceptCLIP} \\
                                 & Mean    & Std.\ Dev.\    & Mean    & Std.\ Dev.\    & Mean    & Std.\ Dev.\    \\
    \midrule
    o4-mini                      & \textbf{0.3116} & 0.0631         & 0.2349  & 0.1093         & 0.3385  & 0.1240         \\
    Gemma3 (unfine-tuned)        & 0.3090  & 0.1201         & 0.2960  & 0.1222         & 0.4856  & 0.1121         \\
    PaliGemma2 w/LoRA            & 0.2329  & 0.0876         & \textbf{0.3834} & 0.1175 & \textbf{0.5823} & 0.1330 \\
    Gemma3 w/LoRA                & 0.2830  & 0.1296         & 0.3631  & 0.1403         & 0.5227  & 0.1392         \\
    ConceptCLIP-Gemma3 w/LoRA    & 0.2323  & 0.1228         & 0.3680  & 0.1171         & 0.5442  & 0.0914         \\
    \bottomrule
  \end{tabular}
\end{table*}

Since SigLIP2 has not been fine-tuned on medical data, the fine-tuned models score lower with SigLIP2 embeddings, while the unfine-tuned model—which excels at general image description—achieves higher scores. In contrast, with MedCLIP and ConceptCLIP embeddings—both of which have been adapted to medical terminology and angiographic imagery—PaliGemma2 obtains the highest average VLScore.

\subsection{VLM Qualitative Analysis}
Building a clinically useful model requires not only quantitative metrics but also qualitative assessment by experienced physicians. In this study, one cardiologist reviewed the reports generated by the fine-tuned VLMs—each trained on the Japanese side of our English–Japanese parallel CAG dataset—according to the following criteria:
\begin{itemize}
  \item Overall score (out of 10)
  \item Laterality errors
  \item Vessel numbering errors
  \item Treatment plan errors
  \item Logical consistency errors
  \item Stenosis detection errors
\end{itemize}

\begin{figure}[t]
  \centering
  \includegraphics[width=\linewidth]{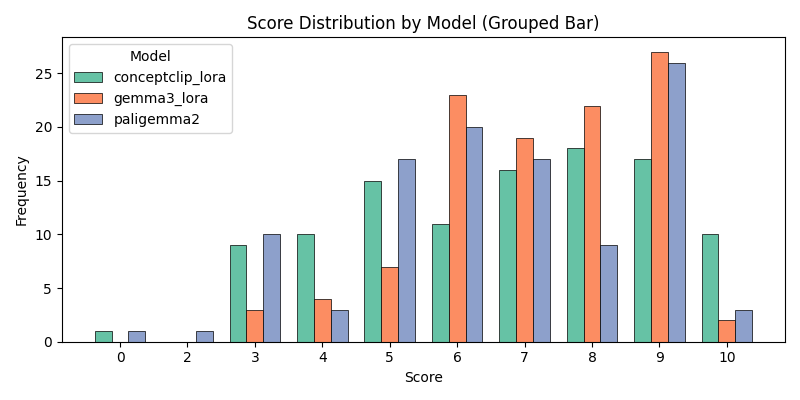}
  \caption{Distribution of physician overall scores (0–10) for each model over 107 cases.}
  \label{fig:vlm-qualitative-hist}
\end{figure}

Figure~\ref{fig:vlm-qualitative-hist} shows the distribution of overall scores across 107 cases, and Table~\ref{tab:vlm-qualitative-stats} summarizes the error counts for each criterion.

\begin{table*}[t]
  \centering
  \caption{Qualitative evaluation results for generated reports (107 cases).}
  \label{tab:vlm-qualitative-stats}
  \begin{tabular}{lccccccc}
    \toprule
    Model                         & Mean  & SD   &
      \makecell[tc]{Laterality\\Error} &
      \makecell[tc]{Vessel\\Error} &
      \makecell[tc]{Treatment\\Error} &
      \makecell[tc]{Logical\\Error} &
      \makecell[tc]{Stenosis\\Detect Error} \\
    \midrule
    Gemma3 w/LoRA                 & \textbf{7.20}  & 1.62 & \textbf{0}   & 76 & \textbf{24} & \textbf{42} & \textbf{27} \\
    ConceptCLIP-Gemma3 w/LoRA     & 6.70  & 2.22 & \textbf{0}   & 84 & 29 & 45 & 46 \\
    PaliGemma2 w/LoRA             & 6.58  & 2.11 & 3   & \textbf{75} & 50 & 55 & 39 \\
    \bottomrule
  \end{tabular}
\end{table*}

The Gemma3 w/LoRA model achieved the highest average score, with most cases receiving 6–9 points. ConceptCLIP-Gemma3 w/LoRA ranked second (mean 6.70) but showed greater score variability. Although PaliGemma2 w/LoRA performed best on VLScore, it received the lowest physician score (mean 6.58). The main factors were logically inconsistent treatment recommendations and laterality misclassifications, which clinicians judged unhelpful. All models also produced vessel numbering errors in about 70\% of cases; since no dedicated segmentation or stenosis‐detection support was used, accurately identifying vessel labels remains challenging.
\section{Discussion}

\subsection{Key Findings}
Our ConvNeXt‐Base CNN achieved high accuracy in both highlight detection and laterality classification (laterality F1 = 0.959, three‐class key‐frame F1 = 0.925), demonstrating its suitability for automatic key‐frame extraction.

Although PaliGemma2 w/LoRA scored highest quantitatively on VLScore, it frequently recommended “watchful waiting” without any medical therapy—an inappropriate treatment plan given standard PCI‐guided care. In contrast, Gemma3 w/LoRA received the highest clinician score (mean 7.20/10) and emerges as our best‐performing model; hereafter, we refer to this model as \textbf{CAG-VLM}. ConceptCLIP-Gemma3 w/LoRA, despite its strong VLScore performance, often repeated treatment-plan text verbatim and misclassified LCX vs.~LAD in several cases (e.g., \#6, \#13), likely due to insufficient fine-tuning of its Vision Encoder replacement.

This contrast is vividly illustrated in Figure~\ref{fig:vlm-output-example}, where PaliGemma2 w/LoRA incorrectly advises no therapy, while ConceptCLIP-Gemma3 w/LoRA correctly identifies LCX \#11 stenosis and frames an appropriate recommendation.

\subsection{Clinical Implications}
By automatically extracting laterality-labeled key frames, our CNN can streamline the image-selection step of coronary-angiography review. These curated images could then be incorporated into future structured-reporting workflows, potentially easing both interpretation and documentation. Conventional visual assessment of invasive angiography still shows about 22 \% inter-observer variability among experienced cardiologists \cite{Shivaie2024Interobserver}. Structured reporting, endorsed by the ACC/AHA/SCAI catheterization-laboratory statement \cite{Sanborn2014-nk}, mitigates such variability; in coronary CT angiography (CCTA), replacing free-form impressions with a structured template raised agreement on the number of significantly stenotic vessels from 53 \% to 68 \% \cite{Ghoshhajra2013-zp}. Although CCTA is non-invasive, its reporting challenges resemble those of cine-CAG, suggesting that similar gains could be realised once structured templates are fully adopted.

CAG-VLM pushes this progress further by coupling CNN-selected frames with a Gemma-3 w/LoRA vision-language model that drafts the narrative report and could substantially shorten report-generation time. In an initial single-reader assessment of 107 cases, the drafts achieved a mean overall rating of 7.2 / 10 (IQR 6 – 9), indicating a promising baseline quality. In a comparable deep-learning CCTA workflow, clinician-supervised AI assistance reduced reading time to about 3.7 minutes per case (a 53–59 \% saving) and improved diagnostic classification, achieving a net reclassification improvement of 0.085, with the largest gains observed in junior readers \cite{Liu2021-og}. Although this benchmark comes from CCTA, similar time-saving and accuracy benefits could potentially emerge once comparable automation is applied to cine-CAG workflows.

Given a residual vessel-numbering error rate of about 70 \% in our current drafts, mandatory expert sign-off remains essential. Taken together, these results suggest that CAG-VLM can accelerate cath-lab documentation, promote consistent terminology, and allow specialists to focus on treatment strategy rather than paperwork—once validated in prospective multi-reader studies.

\subsection{Comparison with Prior Work}
To our knowledge, this is the first dataset that links coronary angiography (CAG) images with physician-validated diagnostic and treatment summaries, enabling multimodal vision-language model (VLM) fine-tuning. While previous CAG datasets such as CathAI and ARCADE concentrate on stenosis detection or segmentation pipelines, our dataset supports a unified workflow that spans image selection through to full report generation. 

\subsection{Limitations}
Our dataset (243 exams, 1,101 frames) is limited in size and contains many atypical angles and low‐quality DICOMs, which clinicians noted can hinder learning of canonical vessel anatomy. 

Annotation noise from original report errors and ambiguous vessel numbering further degrades performance. ConceptCLIP‐Gemma3 w/LoRA often repeated treatment plan text, and misclassified LCX vs. LAD in several cases (e.g., \#6, \#13), likely due to insufficient fine‐tuning of the replaced Vision Encoder.  
All models exhibited vessel‐numbering errors in ~70 \% of cases; without dedicated segmentation or stenosis detectors, accurate labeling remains challenging.  
Finally, our models operate on single frames, whereas real‐world diagnosis integrates multiple cine sequences; extending to full video analysis is essential.

\subsection{Future Work}
We plan to expand the dataset with both typical and atypical cases across multiple centers and imaging systems to improve generalization. Pretraining or jointly fine‐tuning vessel‐specific segmentation modules may reduce numbering errors. Adapting our pipeline to process entire angiographic videos will better mirror clinical decision‐making. Incorporating retrieval‐augmented generation (RAG) with domain guidelines could enhance logical consistency in treatment recommendations. Interactive, clinician‐in‐the‐loop interfaces may further accelerate adoption in practice.
Our work demonstrates that specialist VLMs can acquire complex medical modalities via fine‐tuning and lays the groundwork for automated CAG report generation and treatment planning support.

{
    \small
    \bibliographystyle{ieeenat_fullname}
    \bibliography{main}
}


\end{document}